\documentclass[twoside,leqno,twocolumn]{article}
\usepackage{ltexpprt}

\usepackage{times}
\usepackage{helvet}
\usepackage{courier}
\usepackage{amsmath}
\usepackage{amssymb}
\usepackage{times}
\usepackage{algorithm}
\usepackage{algorithmicx}
\usepackage{algpseudocode}
\usepackage{multicol, blindtext}
\usepackage{graphicx}  
\usepackage{subcaption}
\usepackage{color}
\usepackage{breqn}

\begin{document}

\title{\Large Graph Embedding   with Rich  Information through  Heterogeneous Network} 
\author{ 
   \large Guolei Sun \\[-3pt]
   \normalsize King Abdullah University of Science and Technology  \\[-3pt]
    \normalsize	guolei.sun@kaust.edu.sa \\[-3pt]
  \and
   \large Xiangliang Zhang \\[-3pt]
   \normalsize King Abdullah University of Science and Technology  \\[-3pt]
    \normalsize	xiangliang.zhang@kaust.edu.sa \\[-3pt]
}
\date{}

\maketitle







\begin{abstract} \small\baselineskip=9pt 
Graph embedding has attracted increasing attention due to  its critical application in social network analysis. Most existing algorithms for graph embedding only rely on the typology information and fail to use the copious information in nodes as well as edges. As a result, their performance for many tasks may not be satisfactory. In this paper, we proposed a novel and general framework of representation learning for graph with rich text information through constructing a bipartite heterogeneous network. Specially, we designed a biased random walk to explore the constructed heterogeneous network with  the notion of flexible neighborhood. The efficacy of our method is demonstrated by extensive comparison experiments with several baselines  on various datasets. 
It improves the Micro-F1 and Macro-F1 of node classification by 10\% and 7\% on Cora dataset.
\end{abstract}

\section{Introduction}
Graph embedding, aiming to learn low-dimensional   representations for   nodes in graphs, has attracted a lot of attention recently due to its success in network learning tasks such as node classification~\cite{citeseer,jian2017toward}, link prediction~\cite{pachev2017fast}, and community detection~\cite{community_dection1,community_dection2}.
Inspired by neural language models~\cite{negative_sampling},
Deepwalk is proposed as an unsupervised approach that learns node representations from   network topology~\cite{deepwalk}.
The general idea is to maximize the co-occurrence probability of a target node and its context nodes, which are usually generated by short random walks in networks.
LINE \cite{line} improved Deepwalk by solving large-scare network embedding problem. Node2vec \cite{node2vec} improved Deepwalk by preserving high-order proximity between nodes and introducing more flexible neighborhood for context nodes. 

\begin{figure}
\begin{center}
\includegraphics[width=0.5\textwidth,height=\textheight, keepaspectratio]{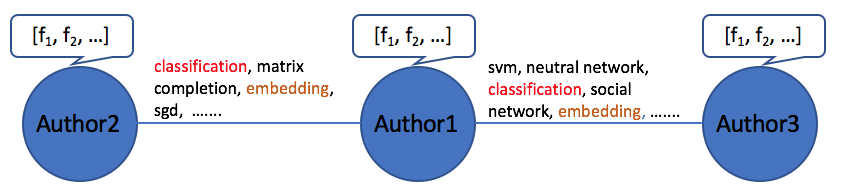}
\end{center}
    \caption{Example of a co-author network} \label{co_author network}
\end{figure}

There is a new trend to integrate multiple types of input information including network topology and node content~\cite{TADW},   neighbors homophily~\cite{zhang2016homophily}, or node labels for designing semi-supervised graph embedding approaches ~\cite{Yang2016RevisitingSL,Pan:2016,tu2016max,li2016discriminative}.
In reality,  networks are complex in terms that not only nodes but also edges contain rich information. For example, in a coauthor network showed in Figure \ref{co_author network}, the nodes  representing authors are associated with a feature set $[f_1, f_2, ...]$, which   contain   information about affiliations, research interests, and working experience. The edges indicating   co-author relationships can be contented by the jointly published papers, which  include key-words such as classification, matrix completion, embedding, social network etc. It is essential that graph embedding should learn from both  \emph{topology information} and  \emph{node/edge content  information}.  

TADW~\cite{TADW} and HSCA~\cite{zhang2016homophily} consider node content information during unsupervised graph representation learning based on matrix decomposition, i.e., decomposing an equivalent word-context matrix converted from network transition matrix, with the help of text information matrix. Other works use deep neural networks to do semi-supervised representation learning, which utilizes text information as  well as label information \cite{Yang2016RevisitingSL}.
Generally, existing approaches have two issues: first, they require matrix operation like SVD decomposition or training of deep neural network, which prohibits them from dealing with large scale graphs; and second, they cannot incorporate the \emph{edge} content information. 
In this paper, we propose a general framework for graph embedding  with rich  information (GERI), which can learn scalable representations for networks with rich text information on nodes and edges. 
GERI is composed of two steps. First, a homogeneous graph with text information on nodes and edges is converted into a heterogeneous graph.
Then, a novel discriminant random-walk method is proposed to preserve the high-order similarity between target nodes, by exploring the network in a mixture of the breadth first search (BFS) and the depth first search (DFS)  manner.

The evaluation of obtained graph embedding is conducted on   multi-label and multi-class classification task with applications to   different publicly available datasets. The results show that GERI outperforms the state-of-the-art TADW by up to 8\% in classification accuracy. GERI is also computationally efficient since its major sections can be easily parallelized. 
The contributions of this paper can be summarized as follows: \\
1. To the best of our knowledge, this is the first work utilizing text information in the homogeneous network through converting into a heterogeneous graph. Homogeneous graph and text information in various forms are integrated into one heterogeneous graph, giving us the possibility to integrate and exploit various information. \\
2.   A biased random walk is proposed to sample appropriate neighbors on the heterogeneous graph for node embedding.  \\
3. Extensive experiments comparing with TADW, Naive combination, node2vec, Line and Deepwalk on various public datasets verify the efficacy of GERI.
\section{Related Work}
\subsection{Homogeneous network embedding}
A great amount of attention has been paid to homogeneous network embedding. The related work can be categorized as \emph{unsupervised} or \emph{semi-supervised} learning approaches depending upon the usage of given labels in the learning process. 
Related works can also be divided by checking if only the network topology information is used for learning, or other information is additionally employed, e.g., node content information.
The study in \emph{unsupervised} representation learning with \emph{only the topology} information has a big family of developed approaches~\cite{graph_embed1,graph_embed2,dimen_reduction1,dimen_reduction2}.  The network topology is usually represented by an adjacency matrix, $A_{(|V|*|V|)}$, where $A_{ij}$ is 1 if node $v_i$  and $v_j$ are connected, or is a numeric weight value if the edge connecting $v_i$  and $v_j$  has a weight, otherwise $A_{ij}$ is 0. To obtain node representation in $\Re^d$, dimensionality reduction techniques like singular value decomposition (SVD) or principal component analysis (PCA), and multidimensional scaling (MDS) can be applied on $A$ or graph Laplacian matrix  and Modularity matrix derived from $A$~\cite{Tang:2009, ahmed2013distributed}.  However, the poor scalability of these approaches makes them difficult to be applied to large-scale networks.  

Recently, another stream of work addresses the unsupervised representation learning of nodes in large-scale graph with an inspiration from neural language models (e.g., word2vec in~\cite{negative_sampling, skipgram}).  Deepwalk~\cite{deepwalk} and node2vec~\cite{node2vec} learn node representation that maximizes the co-occurrence probability of a target node and its context nodes, which are generated by short random walks in networks. Based on the similar notion of ``context'', LINE~\cite{line} is proposed to explicitly preserve the first-order and second-order proximity between nodes. Following the proof that the Skip-Gram model in word2vec with negative sampling~\cite{negative_sampling} is implicitly factorizing a word-context matrix in~\cite{Levy:2014}, Yang et al. show that Deepwalk is equivalent to matrix factorization also when the factorized matrix is designed to show the probability that a node reaches another by random walks~\cite{TADW}. 

Deep learning has been recognized as a promising solution to problems  across several research fields~\cite{Goodfellow-et-al-2016}.
Deep learning also played in network embedding. Auto-encoder is used in~\cite{Wang:2016-kdd} with first-order and second-order proximity for designing a semi-supervised deep learning model for network embedding. Auto-encoder is also employed in~\cite{cao2016deep}  for learning a non-linear mapping from vector representations contained by a positive point-wise mutual information (PPMI) matrix to low-dimensional vertex representations. Convolutional Neural Networks helped on learning embedding of heterogeneous network in~\cite{Chang:2015-KDD}.
However,  these models focus mainly on learning  from \emph{  topology information}. 

In this set of \emph{semi-supervised} representation learning work, available labels guide the learned representation for better predictive capacity.
After the presence of Deepwalk, Yang et al. propose a transductive and inductive framework (Planetoid) for learning the representation for each graph node to jointly predict the class label and the neighborhood context in the graph~\cite{Yang2016RevisitingSL}. A method called max-margin DeepWalk 
proposed in~\cite{tu2016max}  jointly optimizes the max-margin classifier and the representation learning model formulated in a matrix factorization manner. Similarly, Discriminative Deep Random Walk 
proposed in~\cite{li2016discriminative} jointly learns a classifier and vertex representation by combining the loss function of SVM and that of Skip-gram model.

Also many of these unsupervised and semi-supervised  approaches learn from \emph{only the topology information}, except~\cite{TADW,Pan:2016,zhang2016homophily},  which incorporated with node content information and other available graph information.
However, methods like TADW~\cite{TADW} and HSCA~\cite{zhang2016homophily} require matrix operation like SVD decomposition, which prohibits them from dealing with large scale graphs.  TriDNR in~\cite{Pan:2016} is a semi-supervised method and requires extra label information. 
All the above-mentioned approaches are not able to incorporate information on edges, which can be integrated by our proposed model.   

\subsection{Heterogeneous network Embedding}
Heterogeneous network differs from homogeneous network in that the former has more than one type of nodes and edges while the later only has one type of nodes and edge. For example, in Figure \ref{hete_network}, the left is a homogeneous network, where  nodes are authors and   edges  are "author-author" relationships. In contrast, the right   is a heterogeneous network, which contains two types of nodes and two types of edges. Actually, homogeneous network can be considered as a special case  of heterogeneous counterparts.


There are several proposed algorithms targeting on heterogeneous network embedding problem. Embedding of embedding~\cite{eoe_hete} focuses on the embeddings of coupled heterogeneous networks, which is comprised of two different but related sub-networks that are connected by inter-network edges. However, this model cannot solve other type heterogeneous network problems.
Predictive text embedding through large-scale heterogeneous text networks~\cite{pte_text} proposes to embed words under the supervision of labels, by constructing heterogeneous text network.  
It samples the same amount of edges in different types. Even though the algorithm can be extended to other cases, its use is limited because the model only considers low-order proximity (first and second) between nodes.
Context-aware network embedding (CANE)~\cite{cane} is a recently proposed algorithm to embed networks with different contexts related to the nodes. The embeddings for a vertex is the concatenation of two types of embeddings proposed in the paper: structure-based embedding and text-based embedding. However, the paper fails to consider the inter-relation between structure-base and text-based embeddings, which are learned independently. What's more, the model is not general since it cannot be used when rich text information is available on edges.

To sum up, to the best of our knowledge, our method for graph embedding through bipartite heterogeneous network is novel and can be used in more general cases.

\section{Problem Formulation}
Formally, 
let $G=(V,E,T_V,T_E)$ denote a network with rich content information for nodes and edges.
More specifically, $V=\{v_1,v_2,...,v_{|V|}\}$ is a set of nodes,  and $E=\{e=(v_i,v_j):v_i\in V, v_j\in V\}$ is a set of edges linking two nodes.  
$T_V$ is  the word occurrence matrix for nodes, where  each entry  $T_V(i,k)$ indicates the occurrence of word $w_k$ associating with node $v_i$,  and $T_V(i,k)=0$ for the absence of $w_k$ in $v_i$ content.   Similarly, $T_E$ is the  word occurrence matrix for edges, where each entry $T_E(i,j,k) $ indicates the occurrence of word $w_k$ on edge connecting  node $v_i$ and $v_j$, and $T_E(i,j,k) =0$ for the absence.  
The purpose of our work is to learn a low-dimensional representation vector $\textbf{v} \in \Re^d$ for each node $v \in V$, by considering the network topology and rich text information on nodes ($T_V$) and edges ($T_E$).

To incorporate with the text information on nodes and edges, we propose to first convert the homogeneous network into a heterogeneous one, and then learn the representations. \\ 
\textbf{Definition 1: Target nodes \& Bridge nodes}: \emph{Target} nodes are the nodes $V$ in the original homogeneous network $G$, for which embeddings will be learned. When converting $G$ into a heterogeneous one,  \emph{bridge} nodes are created to incorporate the text information, for assisting the embedding learning of target nodes. As shown in Figure \ref{hete_network}, target nodes are in blue, while bridge nodes are in red. The details of bridge nodes construction will be introduced in section \ref{sec:heter}.\\
\textbf{Definition 2: Bipartite heterogeneous network}: A bipartite heterogeneous network is the network whose nodes can be divided into two disjoint groups such that every edge links a node in one group and another node in the second group. In our setting,  one group contains only target nodes, while the other group contains only bridge nodes. 

\section{Method}
In this section, 
we first introduce how to construct the bipartite heterogeneous network for the given problem. Then, we present the objective function to minimize. Finally, we introduce our sampling strategy and overall algorithm.
\subsection{Bipartite Heterogeneous Network Construction} \label{sec:heter}
Given an information network $G=(V,E,T_V,T_E)$, we construct a bipartite heterogeneous network $(V,U,E_{he})$, where $V$ includes the target nodes, $U$ contains the bridge nodes,  $E_{he}$ are the edges between target nodes and bridge nodes.
Bridge nodes are the set of words, $U=\{w_1,w_2,...,w_{|U|}\}$, existing in node and edge text information. 

An edge  in $E_{he}$ connects a target node $v_i$ and a bridge node $w_k$ under two circumstances:\\
1) $v_i$ and $w_k$ are connected when $T_V(i,k) \ne 0$. That is to say, a target node $v_i$ is connected with a bridge node $w_k$ if word $w_k$ occurs in the content information of node $v_i$. The weight associating with the edge is the value of $T_V(i,k)$.\\
2) $v_i$ and $v_j$ are both connected to $w_j$, when $T_E(i,j,k)  \ne 0$. In other words,  target node $v_i$ and $v_j$ are both connected with a bridge node $w_k$ if word $w_k$ occurs in the content information of the edge connecting $v_i$ and $v_j$. 
\begin{figure}
\begin{center}
\includegraphics[width=0.5\textwidth,height=\textheight,keepaspectratio]{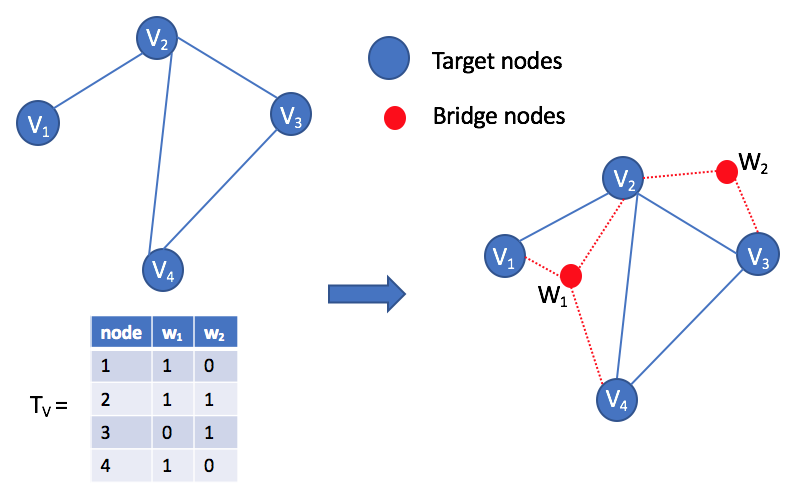}
\end{center}
    \caption{Example of converting a  homogeneous network (left) to a heterogeneous network (right) with bridge nodes.} \label{hete_network}
\end{figure}

Then, the network our algorithm works on is $G_{he}=(V,U,E_{he},E)$, including $E_{he}$ and original $E$, as shown in Figure  \ref{hete_network}.
One   prominent advantage   of using the constructed heterogeneous network is that  the text information is integrated seamlessly with the original network topology. There is no loss of information.

\subsection{Objective Function}
Given $G_{he}=(V,U,E_{he},E)$, our goal is to learn a mapping $f$ such that $f(v_i) \in \Re^d$ is the embedding vector for target node $v_i$, where $d$ is the dimension of our feature representation. Besides,  a bridge node can be also mapped as a feature vector $g(w_k) \in \Re^d$.
Inspired by node2vec, which    maximizes the log-probability of observing network neighborhoods for all the nodes conditioned on their feature representation \cite{node2vec}, we formulate our problem  in heterogeneous network as a maximum likelihood optimization problem with objective function defined as 
{\small
\begin{align*}
	\max_{f,g}\sum_{u\in V}\log(P_r(N_s(u)|f(u)))+\lambda_1 \sum_{u\in U}\log(P_r(N_s(u)|g(u)))
\end{align*}
}%
where $N_s(u)$ is the neighborhood of node $u$.
The first part of the objective is the log-probability of observing network neighborhoods for target nodes, while the second part is the log-probability of observing network neighborhoods for the bridge nodes.   $\lambda_1$ is the balance parameter.

Following the two assumptions made in Deepwalk and node2vec: observing different neighborhoods is independent and feature space is symmetry, we can make our optimization problem solvable and  have the following formulations:
{\small
\begin{align*}
P_r(N_s(u)|f(u))=\prod_{v\in N_S(u)} P_r(v|f(u))\\
P_r(N_s(u)|g(u))=\prod_{v\in N_S(u)} P_r(v|g(u))
\end{align*}
}%
and
{\small
$$
P_r(v|f(u))=\begin{cases}
			\frac{\exp(g(v)\cdot f(u))}{Y(f(u))}&v\in U \\
            \frac{\exp(f(v)\cdot f(u))}{Y(f(u))}&v\in V\\
		 \end{cases} 
$$  
$$
P_r(v|g(u))=\begin{cases}
			\frac{\exp(g(v)\cdot g(u))}{Y(g(u))}&v\in U \\
            \frac{\exp(f(v)\cdot g(u))}{Y(g(u))}&v\in V\\
		 \end{cases} 
$$ 
}%
where,
{\small
\begin{align*}
Y(f(u))=\sum_{v\in U}\exp(g(v)\cdot f(u))+\sum_{v\in V}\exp(f(v)\cdot f(u))\\
Y(g(u))=\sum_{v\in U}\exp(g(v)\cdot g(u))+\sum_{v\in V}\exp(f(v)\cdot g(u))
\end{align*}
}%
Using the above formula, the objective is written as follows:
{\small
\begin{dmath*}
\max_{f,g}\sum_{u\in V}\left[-\log Y(f(u))+\sum_{\substack{n_i\in N_s(u) \\ n_i\in U}}{g(n_i)\cdot f(u)} +\sum_{\substack{n_i\in N_s(u) \\ n_i\in V}}f(n_i)\cdot f(u) \right]+\lambda_1 \sum_{u\in U}\left[-\log Y(g(u))+\sum_{\substack{n_i\in N_s(u) \\ n_i\in V}}f(n_i)\cdot g(u)+\sum_{\substack{n_i\in N_s(u) \\ n_i\in U}}g(n_i)\cdot g(u)\right]
\end{dmath*}
}%
While $Y(f(u))$ and $Y(g(u))$ can be approximated by negative sampling, $N_s(u)$ has a significant influence on the optimization results.
Inspired by  node2vec, which proposed a concept of flexible neighborhood in homogeneous network, we propose a novel randomized procedure that can sample neighborhood of a source node in heterogeneous network.

\subsection{Local search: BFS \& DFS}
Node2vec treats the problem of sampling neighborhoods of a source node in homogeneous network as a form of local search. Similarly, we use the local search to deal with the problem of sampling neighborhoods in heterogeneous network. We propose a novel sampling method, which can explore the heterogeneous graph in a more elegant and controlled manner such that better neighbors of nodes can be obtained.
We first discuss the two  extreme sampling strategies, BFS and DFS, before introducing our proposed method. \\
\begin{figure}
\begin{center}
\includegraphics[width=0.5\textwidth,height=\textheight,keepaspectratio]{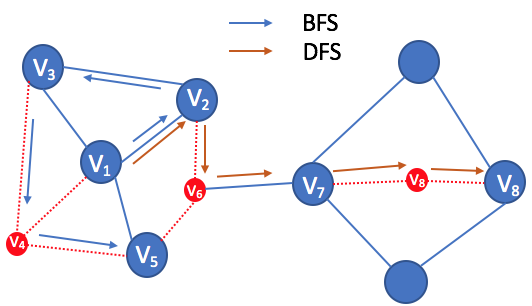}
\end{center}
    \caption{Two extreme search strategies in heterogeneous network: BFS and DFS} \label{bfs_dfs}
\end{figure}
\textbf{Breadth first search (BFS):} Starting with the source node $v_1$, this strategy samples the nodes that are within a certain small distance from $V_1$. In Figure \ref{bfs_dfs}, the sampling of BFS starting from source node $V_1$ can be $V_2,V_3$ etc. This strategy can preserve the community and cluster information in the graph. Those nodes that belong to the same cluster will have similar embeddings. \\
\textbf{Depth first search (DFS):} Starting with the source node $V_1$, this strategy explores the graph as deep as possible. In  Figure \ref{bfs_dfs}, DFS starting at $V_1$ can visit nodes sequence: $V_2, V_6, V_7,...$. This strategy can preserve the structure information such that those nodes that do not belong to the same cluster, but have the same structure around it, will have similar embeddings.

\subsection{Sampling strategy}
\begin{figure}
\begin{center}
\includegraphics[width=0.5\textwidth,height=\textheight,keepaspectratio]{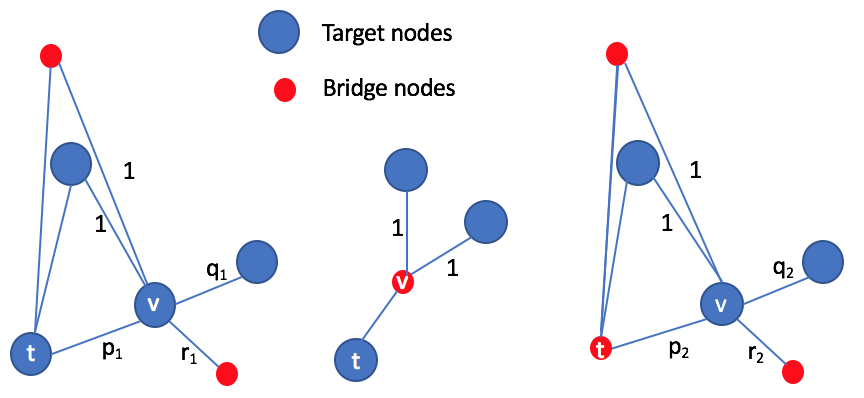}
\end{center}
    \caption{Three  cases of random walk  in heterogeneous network, giving that random walk just visited $v$ from $t$: left (both $t$ and $v$ are target nodes); middle ($t$ is target node and $v$ is bridge node); right ($t$ is bridge node and $v$ is target node).} \label{search_bias}
\end{figure}

We design a compromised search paradigm between BFS and DFS for our heterogeneous network, because the search method in node2vec is designed for homogeneous network. The existing sampling strategy in PTE~\cite{pte_text} can only preserve the low proximity between nodes, which is not usually desirable.
Consider a random walk that just walked from node $t$ to  $v$ in Figure \ref{search_bias}. Then it needs to decide where to go in the next step, which depends on the transition probability $\beta_{vx}$ between node $v$ and next node $x$, and the types of previously visited node $v$ and $t$.

We define   the transition probability $\beta_{vx}$ in three cases:\\
{\bf Case 1: } node $t$ and $v$ are both \emph{target} nodes, as shown in the left example of Figure \ref{search_bias}. 
The next node to visit from $v$ can be a target node, or a bridge node. We introduce three parameters $p_1$, $q_1$, and $r_1$ to guide the walk, and discuss their meanings later. Given the weight $e_{vx}$ between node $v$ and $x$ (e.g., measured as the occurrence frequency of a word in a node/edge), the transition probabilities $\beta_{vx}$ is defined as: 
$$
\beta_{vx}=\begin{cases}
	   p_1*e_{vx} \quad \text{     if  } d_{tx}=0 \\
            1*e_{vx} \quad \text{     if  } d_{tx}=1 \\
            q_1*e_{vx} \quad \text{     if  } d_{tx}=2 ,  x \in V \text{ target nodes}\\
            r_1*e_{vx}  \quad \text{     if  } d_{tx}=2 , x \in U \text{ bridge nodes}\\         
		 \end{cases} 
$$ 
where $d_{tx}$ denotes the shortest path distance between nodes $t$ and $x$. \\
{\bf Case 2:} node $t$ and $v$ are \emph{target} node and \emph{bridge} node, respectively (the middle example shown in Figure \ref{search_bias}). In this case, we don't allow the walk to go back and  expect the walk to explore more target nodes because we focus more on the relationship between a target node and other target nodes. The transition probabilities $\beta_{vx}$ is defined as: 
$$
\beta_{vx}=\begin{cases}
			0 \quad \quad \quad   \text{     if  } d_{tx}=0 \\
            1*e_{vx}  \quad \text{     if  } d_{tx}\ne 0 \\
		 \end{cases} 
$$  
{\bf Case 3:}  node $t$ and $v$ are \emph{bridge} node and \emph{target} node, respectively (the right example shown in Figure \ref{search_bias}). We introduce three parameters $p_2$, $q_2$, and $r_2$ to guide the walk. The transition probabilities $\beta_{vx}$ is as follows:
$$
\beta_{vx}=\begin{cases}
		p_2*e_{vx} \quad \text{     if  } d_{tx}=0 \\
            1*e_{vx} \quad \text{     if  } d_{tx}=1 \\
            q_2*e_{vx}  \quad \text{     if  } d_{tx}=2, x \in V  \text{ target nodes}\\
            r_2*e_{vx}  \quad \text{     if  } d_{tx}=2,  x \in U  \text{ bridge nodes}\\         
		 \end{cases} 
$$   

In the following, we discuss the meaning of the parameters and their implications.\\
{\bf Back parameter $p$}. Parameter  $p_1$ and $p_2$  control the probability to revisit the node that has been visited in the second last step. Setting it to a small value means that the walk is less likely to go back. However, setting it to a large value ($>1$) means the walk is more likely to visit the local neighbors of the source node. Then, it is more like the BFS search.\\
{\bf Out-target parameter $q$}. Parameter $q_1$ and $q_2$, on the one hand, control the likelihood of visiting target nodes in the random walk. If $q_1$ ($q_2$) is greater than $r_1$ ($r_2$), then the random walk is more likely to visit target nodes, which means target nodes play a more important role in the random walk. On the other hand, $q_1$ and $q_2$ control the depth of  exploring the graph. If $q_1$ and $q_2$ are large, then the random walk is more likely to go as deep as possible, which is like DFS search. However, if $q_1$ and $q_2$ are small, the random walk is more likely to visit the local nodes of the source, which is more like BFS search.\\
{\bf Out-bridge parameter $r$}.   Contrary to $q$,  $r$   controls the likelihood of visiting bridge nodes in the random walk. If $q_1$ ($q_2$) is less than $r_1$ ($r_2$), then the random walk is more likely to visit target nodes. Similar to $q$, $r$  controls the probability to explore the graph deeply. If it are high ($>1$), the walk is more like DFS. Otherwise, the walk is more like BFS.\\
In practice, since each pair has the same meaning, we set $p_1=p_2=p$, $q_1=q_2=q$ and $r_1=r_2=r$, to  largely reduce the burden of parameter altering.  

\begin{algorithm}[htb]
\caption{GERI algorithm}
\label{alg:tahn} 
\begin{algorithmic}[1]
\Require $G=(V,E,T_V,T_E)$, Dimensions $d$, walks per vertex $\gamma$,  window size $\tau$, walk length $l$, balance parameter $\lambda_1$, and $p$, $q$,$r$
\Ensure {matrix of nodes representation $\Theta \in \Re^{|V|*d}$}
\State Initialize $\Theta$ by standard normal distribution
\State Construct $G_{he}=(V,U,E_{he},E)$
\State $\beta$=PreprocessBiasWeight($G_{he},p,q,r$)
  \For {$iter=1$ to $\gamma$}
     \State $\phi$=shuffle($V$)
     \For {all nodes $v \in \phi$}
          \State walk=RandomWalk($G_{he},\beta, v, l$)
          \State trainpairs=GenerateSkipGramTraining(walk, $\tau$)
          \For {$(v_1, v_2) \in$ trainpairs}
              \If {$v_1$ is a target node}
                   \State SGD(k,d,$(v_1, v_2)$,$\eta$)
              \Else
                   \State SGD(k,d,$(v_1, v_2)$,$\lambda_1 \eta$)
              \EndIf
          \EndFor
     \EndFor
   \EndFor \\
\Return $\Theta$
\end{algorithmic}
\end{algorithm}

\subsection{GERI algorithm}
We show the pseudo-code of our method in Algorithm \ref{alg:tahn}. Given $G=(V,E,T_V,T_E)$, the purpose of our algorithm is to learn d-dimensional representation for all nodes in $V$. After constructing a heterogeneous network and defining the transition probabilities $\beta$,  in each iteration, the random walk of length $l$ is done for each node (target node and bridge node), using the sampling strategy introduced above. Then the walk is processed to generate standard SkipGram training pairs. For each training pair, we use Stochastic Gradient Descent~\cite{sgd} to optimize the object function. 
Our algorithm is easily parallelizable and can be executed asynchronously, leading its efficacy and scalability.

\section{Experiments}

We evaluated our algorithm on various datasets including citation network and language network on the task of multi-class classification and multi-label classification, and compared it with other state-of-art graph embedding algorithms.
\begin{table*}[t]
\scriptsize
\centering
\caption{Comparison of Micro-F1 and Macro-F1 score on Citeseer datasets for different dimensions}
\label{result_cite}
\begin{tabular}{|c|c|c|c|c||c|c|c|c|c|}
\hline
\multicolumn{5}{|c||}{\texttt{Micro-F1}} & \multicolumn{5}{|c|}{\texttt{Macro-F1}} \\
\hline
Algorithm & d = 16 & d = 32 & d = 64 & d=128 & Algorithm & d = 16 & d = 32 & d = 64 & d=128 \\ \hline
\textbf{Deepwalk} & 0.5306	&0.546&	0.5832&	0.5832 &
 \textbf{Deepwalk} &0.4755	&0.4981&	0.4981&	0.5359
 \\ \hline
\textbf{Line} & 0.4695&	0.6757&	0.6908	&0.7044&
 \textbf{Line} & 0.4695&	0.6112	&0.6355&	0.6557
\\ \hline
\textbf{Node2vec} & 0.5380	&0.5524	&0.5692&	0.6029&
\textbf{Node2vec} & 0.4633	&0.4828&	0.5108	&0.5539
 \\ \hline
 \textbf{Text Only} & 0.7030	&0.7131	&0.7121	&0.7039
& \textbf{Text Only} &0.6305&0.6559	&0.6683	&0.6677
  \\ \hline
\textbf{Naive combination} & 0.7070&	0.7042&	0.7138&	0.7012
 & \textbf{Naive combination} &0.6684	&0.6665	&0.6779&	0.6664
 \\ \hline
\textbf{TADW} & 0.3952	&0.5139&	0.6751&	0.718
& \textbf{TADW} & 0.3121	&0.4465&	0.6156&	0.6771
 \\ \hline
\textbf{Deepwalk(hete)} & 0.7077&0.7114	&0.7213&	0.6993
& \textbf{Deepwalk(hete)} &0.6332&	0.6531	&0.6725	&0.6576
 \\ \hline
\textbf{Line(hete)} &  0.7159&	0.7228	&0.7165	&0.7065
 & \textbf{Line(hete)} & 0.6408	&0.6569&	0.655&	0.6571
 \\ \hline
 \textbf{Node2vec(hete)} & 0.7185	&0.7219	&0.7222	&0.7077
 & \textbf{Node2vec(hete)} & 0.6487	&0.6645&	0.6764	&0.6742
 \\ \hline
\textbf{GERI} & \textbf{0.7427}	&\textbf{0.7472}&	\textbf{0.7457}	&\textbf{0.7358}
 & \textbf{GERI} & \textbf{0.6794}&	\textbf{0.6962}	&\textbf{0.7004}	&\textbf{0.6944}
 \\ \hline
\end{tabular}
\end{table*}

\begin{table*}[t]
\scriptsize
\centering
\caption{Comparison of  Micro-F1 and Macro-F1 score on Cora datasets for different dimensions}
\label{result_cora}
\begin{tabular}{|c|c|c|c|c||c|c|c|c|c|}
\hline
\multicolumn{5}{|c||}{\texttt{Micro-F1}} & \multicolumn{5}{|c|}{\texttt{Macro-F1}} \\
\hline
Algorithm & d = 16 & d = 32 & d = 64 & d=128 & Algorithm & d = 16 & d = 32 & d = 64 & d=128 \\ \hline
\textbf{Deepwalk} & 0.7569	&0.7757	&0.8013&	0.8151
 &\textbf{Deepwalk} &0.7421	&0.7645	&0.7917&	0.8041
 \\ \hline
\textbf{Line} & 0.7323	&0.7179	&0.7090&	0.7127 &\textbf{Line} & 0.7142	&0.7080	&0.7048	&0.7045
\\ \hline
\textbf{Node2vec} & 0.7762	&0.7936&	0.8096&	0.8206
&
\textbf{Node2vec} & 0.7651	&0.7829	&0.8&	0.811
 \\ \hline
 \textbf{Text Only} &0.7242	&0.7399	&0.7344	&0.6957
& \textbf{Text Only} &0.6989&0.718	&0.7097&	0.6651
 \\ \hline
\textbf{Naive combination} & 0.7864&	0.8070&	0.8198	&0.8148
& \textbf{Naive combination} &0.7629	&0.7898	&0.8033&	0.8012
\\ \hline
\textbf{TADW} & 0.6732&	0.7736	&0.825	&0.8279
 & \textbf{TADW} & 0.5676	&0.7400	&0.808&	0.8093
 \\ \hline
\textbf{Deepwalk(hete)} & 0.7858	&0.8065&	0.7951	&0.7962
& \textbf{Deepwalk(hete)} &0.7648	&0.7867&	0.7757	&0.7790
 \\ \hline
\textbf{Line(hete)} & 0.7928	&0.8131&	0.7903&	0.7866
 & \textbf{Line(hete)} & 0.7928&	0.7927&	0.7663	&0.7703
 \\ \hline
 \textbf{Node2vec(hete)} & 0.8172&	0.8131&	0.8064	&0.7920
 & \textbf{Node2vec(hete)} & 0.7957&	0.7948&	0.7836&	0.7689
 \\ \hline
\textbf{GERI} & \textbf{0.8559}&	\textbf{0.8641}	&\textbf{0.8700}	&\textbf{0.8637}
 & \textbf{GERI} & \textbf{0.8423}&	\textbf{0.8508}&	\textbf{0.8582}&	\textbf{0.8509}
 \\ \hline
\end{tabular}
\end{table*}

\begin{table*}[t]
\scriptsize
\centering
\caption{Comparison of  Micro-F1 and Macro-F1 score on DBLP datasets for different dimensions}
\label{result_dblp}
\begin{tabular}{|c|c|c|c|c||c|c|c|c|c|}
\hline
\multicolumn{5}{|c||}{\texttt{Micro-F1}} & \multicolumn{5}{|c|}{\texttt{Macro-F1}} \\
\hline
Algorithm & d = 16 & d = 32 & d = 64 & d=128 & Algorithm & d = 16 & d = 32 & d = 64 & d=128 \\ \hline
\textbf{Deepwalk} & 0.5600&	0.5769	&0.5839&	0.6027 &
 \textbf{Deepwalk} &0.4552&	0.4896&	0.5114	&0.5386
\\ \hline
\textbf{Line} & 0.5220	&0.4939	&0.4895&	0.5080&
 \textbf{Line} &0.4193	&0.3920	&0.3946	&0.4291
\\ \hline
\textbf{Node2vec} &0.5760&	0.5860	&0.5952	&0.6112&
\textbf{Node2vec} &0.4858&	0.5040&	0.525& 0.5466
 \\ \hline
\textbf{Text Only} &0.6113&	0.6472	&0.6698	&0.6894&
\textbf{Text Only} &0.6044	&0.6333	&0.6521	&0.6721
 \\ \hline
\textbf{Naive combination} &0.7440&	0.7476	&0.7524	&0.7511
 & \textbf{Naive combination} &0.718&	0.7233&	0.7284&	0.7300 \\ \hline
\textbf{TADW} &0.5023&	0.6031&	0.6657	&0.7179
& \textbf{TADW} & 0.4925&0.5904	&0.6497&	0.697
\\ \hline
\textbf{Deepwalk(hete)} &0.7555&	0.7582	&0.7684&	0.7771
& \textbf{Deepwalk(hete)} &0.7299&	0.7319&	0.7451	&0.7556
\\ \hline
\textbf{Line(hete)} &0.7669&	0.7703&	0.7792	&0.7853
 & \textbf{Line(hete)} & 0.7442	&0.7479	&0.7578	&0.7648
 \\ \hline
 \textbf{Node2vec(hete)} &0.7553&	0.7623	&0.7716	&0.7787
 & \textbf{Node2vec(hete)} &0.7294&	0.7387	&0.7495	&0.7569
 \\ \hline
\textbf{GERI} & \textbf{0.7695} &	\textbf{0.7786} &	\textbf{0.7867}	& \textbf{0.7919}
 & \textbf{GERI} &\textbf{0.7474}&\textbf{0.7548}	&\textbf{0.7646}	&\textbf{0.7724}
 \\ \hline
\end{tabular}
\end{table*}

\begin{table*}[t]
\scriptsize
\centering
\caption{Comparison of  Micro-F1 and Macro-F1 score on Wiki datasets for different dimensions}
\label{result_wiki}
\begin{tabular}{|c|c|c|c|c||c|c|c|c|c|}
\hline
\multicolumn{5}{|c||}{\texttt{Micro-F1}} & \multicolumn{5}{|c|}{\texttt{Macro-F1}} \\
\hline
Algorithm & d = 16 & d = 32 & d = 64 & d=128 & Algorithm & d = 16 & d = 32 & d = 64 & d=128 \\ \hline
\textbf{Deepwalk} & 0.6125&	0.6235&	0.6354	&0.6434 &
 \textbf{Deepwalk} &0.4559&	0.4837&	0.5158&	0.5354
 \\ \hline
\textbf{Line} & 0.5623	&0.6023&	0.6174&	0.6146&
 \textbf{Line} & 0.3870	&0.4632	&0.4793	&0.4891
\\ \hline
\textbf{Node2vec} & 0.6028&	0.6241	&0.6402	&0.6456&
\textbf{Node2vec} & 0.4307&	0.4885&	0.5007	&0.5147
 \\ \hline
 \textbf{Text only} & 0.7190&	0.7489&	0.7485	&0.6902
& \textbf{Text Only} &0.6165&	0.6462&	0.6403&	0.6735
\\ \hline
\textbf{Naive combination} & 0.7439&	0.7822	&0.7963	&0.8013
& \textbf{Naive combination} &0.6280	&0.6875	&0.6959	&0.7070
 \\ \hline
\textbf{TADW} & 0.3954&	0.5601&	0.7385	&0.7592
& \textbf{TADW} &0.2467	&0.4472	&0.6313	&0.6303
 \\ \hline
\textbf{Deepwalk(hete)} &0.7587&	0.7865&	0.7914	&0.7775
& \textbf{Deepwalk(hete)} &0.6485&	0.6930&	0.6876	&0.6785
 \\ \hline
\textbf{Line(hete)} & 0.7381	&0.7685&	0.7928	&0.7762
 & \textbf{Line(hete)} & 0.5673&	0.6625&	0.6856	&0.6768
 \\ \hline
 \textbf{Node2vec(hete)} & 0.7707	&0.7945	&0.7993&	0.7904
 & \textbf{Node2vec(hete)} &0.6579&	0.6794	&0.6825	&0.6786
 \\ \hline
\textbf{GERI} & \textbf{0.7755}&	\textbf{0.7990}	&\textbf{0.8090}	&\textbf{0.8129} & \textbf{GERI} & \textbf{0.6666}	&\textbf{0.7000}	&\textbf{0.7026}	&\textbf{0.7144}
 \\ \hline
\end{tabular}
\end{table*}

\begin{figure*}[t!]
    \centering
    \begin{subfigure}[t]{0.5\textwidth}
        \centering
        \includegraphics[width=0.75\textwidth]{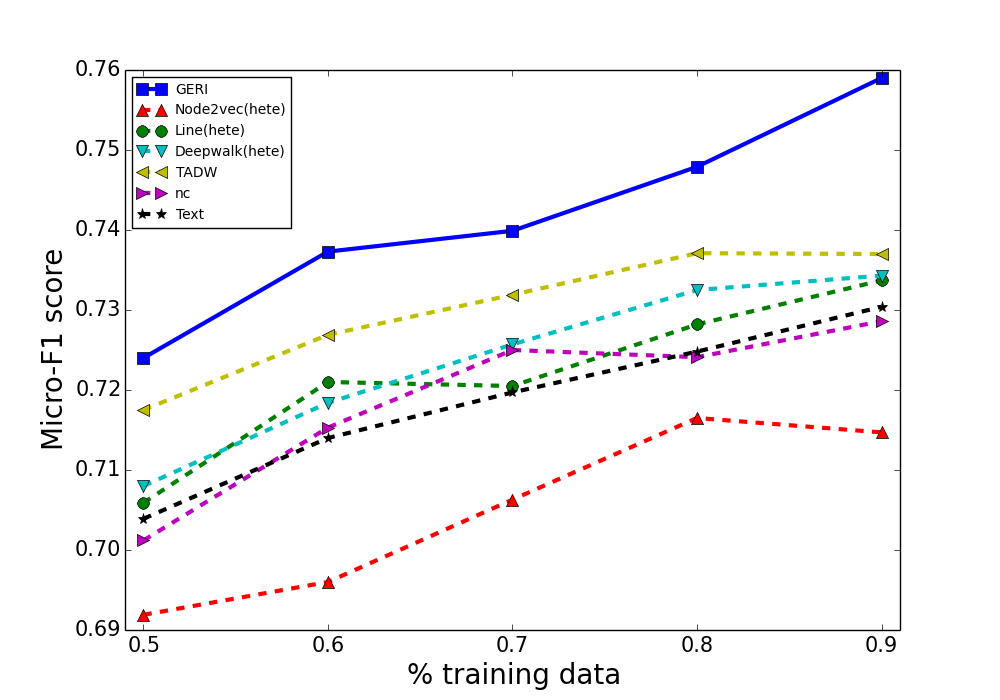}
        \caption{Micro-F1 score w.r.t. \% training data}
    \end{subfigure}%
    ~ 
    \begin{subfigure}[t]{0.5\textwidth}
        \centering
        \includegraphics[width=0.75\textwidth]{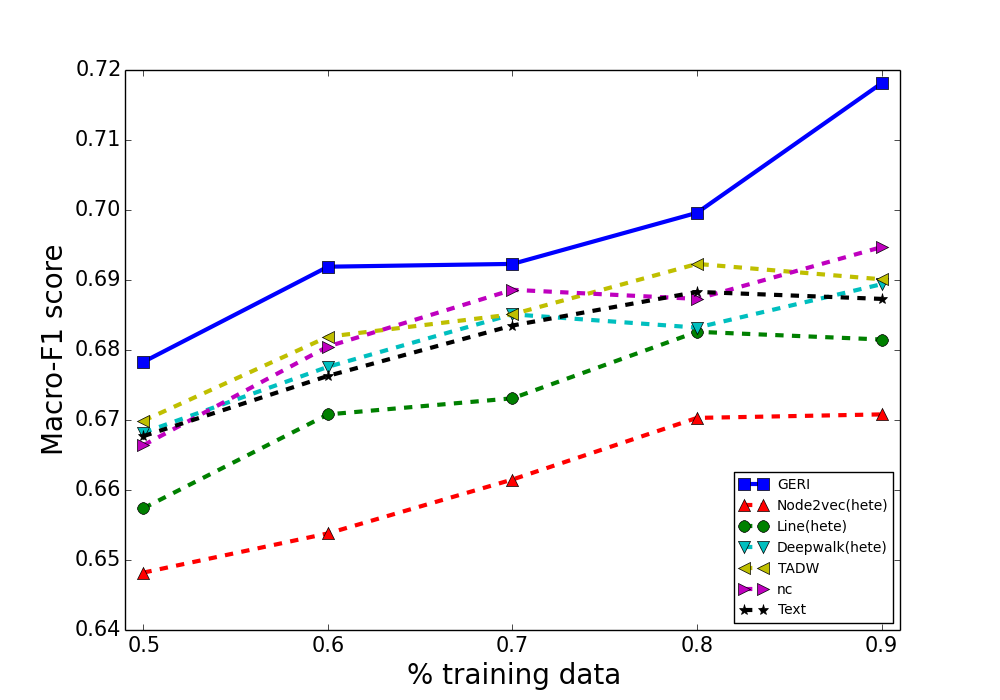}
        \caption{Macro-F1 score w.r.t. \% training data}
    \end{subfigure}
    \caption{Performance comparison w.r.t. different percentage of training data on Citeseer}\label{per_cite}
\end{figure*}

\subsection{Experiment Setup}
In this part, we introduce the datasets, and baselines we used.
\subsubsection{Dataset}
We employ four benchmark networks with text information on nodes, though our approach can work on networks with information on nodes and edges (datasets with information on edges are less accessible than that with information on nodes).\\
\textbf{(1) Citeseer:} it is a citation network, containing 3312 papers from 6 classes and 4732 links between them. The links represent the citation relationships. Each paper has a binary 1433-dimension feature, representing the presence of 1433 important words in the paper. This data set and the next Cora and Wikipedia data set were used for the evaluation of TADW in~\cite{TADW}. We thus also employ the same data sets in this paper.\\
\textbf{(2) Cora:} it is a citation network, containing 2708 publications from 7 classes and 5429 links. The links are citation relationship between documents. Each publication is described by a binary feature of 1433 dimension, indicating the presence of corresponding words.  \\
\textbf{(3) Wikipedia:} it is the subgraph of the co-occurrence network of words extracted from wikipedia, containing 2405 words and 17981 links. The text features are described by a TF-IDF matrix. We extracted the sub-matrix of TF-IDF matrix to do the experiments.  \\
\textbf{(4) DBLP:}  it is a co-author network containing 28702 authors and 66832 links. A link exists as long as two authors co-published a paper. The weight of edges denotes the number of papers that are published together by the related authors. Each node has some labels out of 4 labels, which represent research areas of the author. Each author is described by a 3000-dimension feature, indicating the presence of corresponding words in the paper~\cite{dblp_dataset}. 
 
\subsection{Comparison Algorithm} 
We compare our proposed method with the most popular graph embedding approaches: Deepwalk, LINE, Node2vec and TADW. Among these baseline approaches, only TADW considers the node content information, while others use only network topology information. To have a fair comparison with them, we also apply Deepwalk, LINE and node2vec on our generated heterogeneous network, for verifying the idea of constructing  heterogeneous network and the proposed biased sampling method in GERI.\\
{\bf Deepwalk}: it learns a d-dimension embedding from homogeneous network through uniform random walk, which can be seen as the special case of Node2vec. We set the length of random walk as 150, the number of walk per node as 10 and the number of negative sampling as 5. \\
{\bf LINE}: this method learns the first d/2-dimension embedding from the first-order proximity and learns the second d/2-dimension embedding from the second-order proximity. Then the embedding is the concatenation of the two learned parts. It is applied to the original homogeneous network. The number of negative sampling is also set as 5. We set the number of sampling as 1000M, to get the best performance of this method.\\
{\bf Node2vec}: this method learns d-dimension embeddings from homogeneous network through biased random walk. We choose the best $p$ and $q$ from the list of [0.25, 0.5, 1, 2, 4] by grid search, and show the best performance.\\
{\bf Text-only}: we also evaluate when only text information is used for node classification. We applied SVD for dimensionality reduction, which can be considered also as a kind of new representation learning. \\
{\bf Naive Combination}: this embedding is a naive combination of node2vec embedding and text-only embedding.\\
{\bf TADW}: it learns d-dimension embeddings from matrix decomposition, considering both network topology and node content.\\ 
{\bf Deepwalk(hete)}: we   feed the constructed heterogeneous network to Deepwalk, for evaluating the proposed heterogeneous network construction, and the biased sampling method.\\
{\bf LINE(hete)}: we  also  apply LINE on the constructed heterogeneous network.\\
{\bf Node2vec(hete)}: we apply node2vec one  the same heterogeneous network.\\
{\bf GERI}: we set the parameters of our proposed method, including the number of walk, the length of walk, the number of walk per node and the number of negative sampling,  the same as Deepwalk and Node2vec, in order to make fair comparisons. The balance coefficient $\lambda_1$ is set as 1 (default). We use grid search to choose best $p$, $q$, and $r$. \\
All the representation vectors are finally normalized by setting its L2-norm as 1. We use logistic classification to evaluate all the embeddings with $C=100$. All results shown in the following are the average of 10 independent experiments.

\subsection{Result}

We report the performance of different algorithms under various embedding dimensions (16, 32, 64, and 128) on  Citeseer, Cora,  DBLP and Wikipedia in Table \ref{result_cite}, \ref{result_cora}, \ref{result_dblp} and \ref{result_wiki}, respectively. We use 50\% of data as training data and another 50\% as testing data. There are several interesting observations: \\
\begin{figure*}[t!]
    \centering
    \begin{subfigure}[t]{0.5\textwidth}
        \centering
        \includegraphics[width=0.75\textwidth]{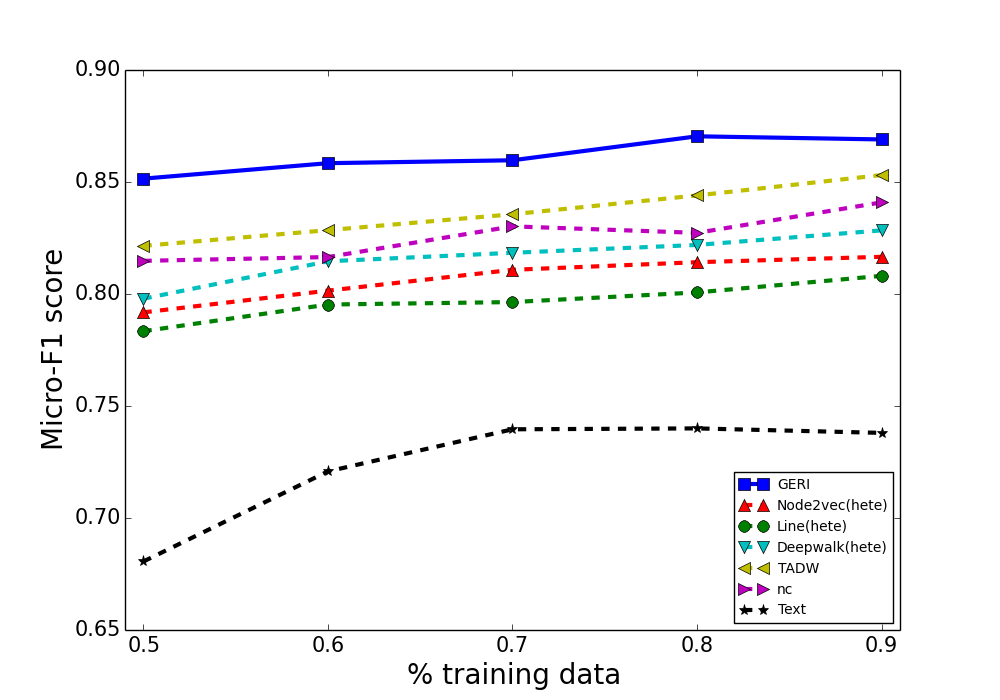}
        \caption{Micro-F1 score w.r.t. \% training data}
    \end{subfigure}%
    ~ 
    \begin{subfigure}[t]{0.5\textwidth}
        \centering
        \includegraphics[width=0.75\textwidth]{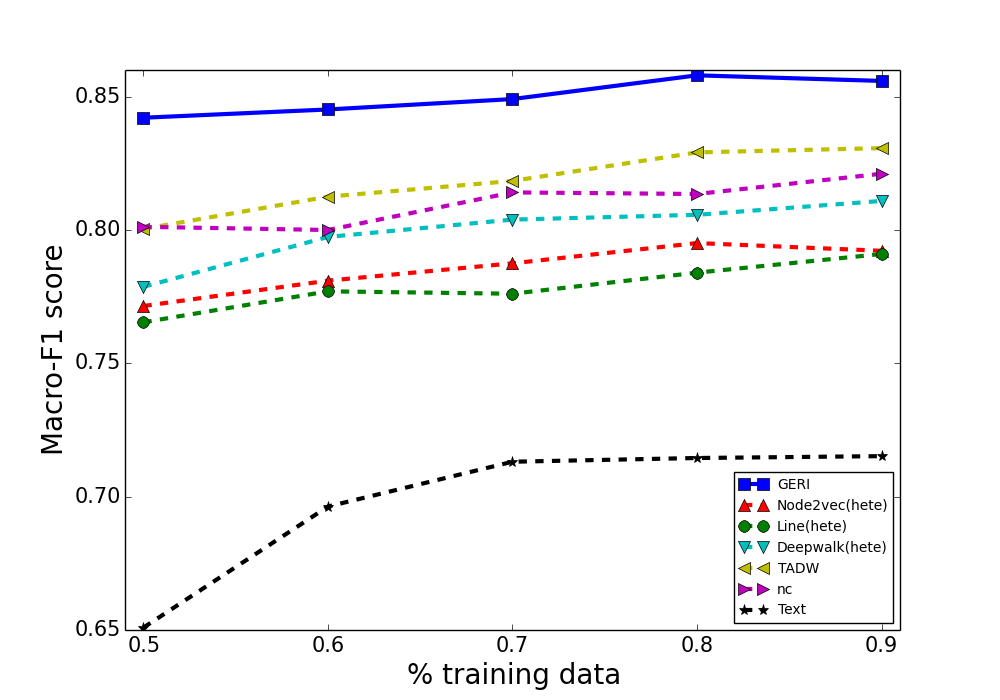}
        \caption{Macro-F1 score w.r.t. \% training data}
    \end{subfigure}
    \caption{Performance comparison w.r.t. different percentage of training data on Cora}\label{per_cora}
\end{figure*}
{\bf 1)} Our proposed algorithm GERI outperforms other algorithms on all data sets with various embedding dimensions. \\
{\bf 2)} Deepwalk(hete), Line(hete) and Node2vec(hete) have better performance than Deepwalk, Line and Node2vec that are applied to the original homogeneous network. It thus verifies that our constructed  heterogeneous network effectively integrates the network topology information and rich text information. Also, our proposed biased sampling method is better than the sampling methods in these network embedding approaches. \\
{\bf 3)} The performance of TADW with low dimensions such as $d=$16, 32 and 64 is worse than others in Citeseer and Cora data set. However,   TADW performs well when $d$ increases to 128. It  is only inferior to GERI, but better than the rest. 
In DBLP and Wikipedia dataset, TADW outperforms Deepwalk, Line and node2vec, but is no better than Deepwalk(hete), Line(hete), and node2vec(hete). Interestingly, the performance of homogeneous algorithms directly feeding with heterogeneous network is excellent, which is also observed in PTE \cite{pte_text}.

To further verify the effectiveness of our algorithm, we conduct experiments on Citeseer and Cora with changing percentage of training data when representation dimension is 128. Results of Deepwalk, Line and Node2vec on homogeneous network are not included because their performance is much worse than other ways. The results for Citeseer and Cora are shown in Figure \ref{per_cite} and \ref{per_cora}, respectively. It shows that our algorithm outperforms other methods in different percentage of training data. 

\begin{figure}
\begin{center}
\includegraphics[width=0.45\textwidth]{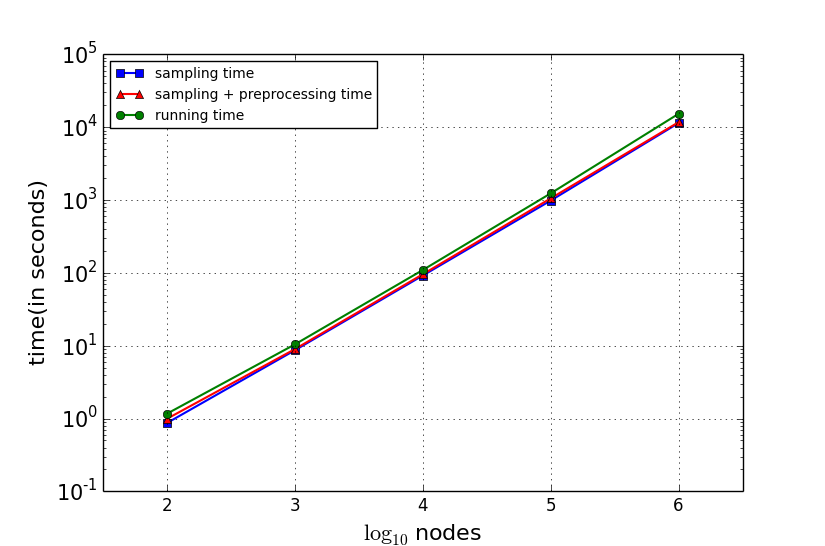}
\end{center} \caption{Scalability of GERI  } \label{run_time}
\end{figure}
\subsection{Scalability}
We test GERI on Erdos-Renyi random graph (degree as 10) with  100, 1000, 10,000, 100,000 and 1,000,000 nodes, for evaluating its scalability. All the results are the average of 10 independent experiments. As shown in Figure \ref{run_time}, the empirical computational time of GERI is almost linear with respect to the number of nodes. It means that GERI if parallelized can be applied on large-scale network embedding.  

\section{Conclusion}
This paper studies the problem of network embedding with rich  information on nodes as well as edges.
A GERI approach is proposed to firstly integrate copious information into a heterogeneous network, and then  learn embeddings for target nodes through the constructed network,  based on the new designed biased random walk. As shown in the experiments, GERI has better performance over several state-of-the-art algorithms on the task of multi-label and multi-class node classification. In the future, we plan to  incorporate the label information to guide the process of embedding learning, i.e., designing a supervised learning approach. It will be also interesting to   extend our proposed biased random walk to various types of heterogeneous network. 


\bibliographystyle{siam}
\bibliography{ref-short}

\end{document}